\documentclass[10pt,twocolumn,letterpaper]{article}

\usepackage[pagenumbers]{cvpr} %

\usepackage{times}
\usepackage{epsfig}
\usepackage{graphicx}
\usepackage{amsmath}
\usepackage{amssymb}
\usepackage{booktabs}
\usepackage[table,xcdraw]{xcolor}
\usepackage{caption}
\usepackage{float}
\usepackage{placeins}
\usepackage{booktabs}
\usepackage{colortbl}
\usepackage{stfloats}
\usepackage{enumitem}
\usepackage{multirow}
\usepackage[table,xcdraw]{xcolor}
\usepackage{multirow}
\usepackage{makecell}

\usepackage[pagebackref,breaklinks,colorlinks]{hyperref}

\usepackage[capitalize]{cleveref}
\crefname{section}{Sec.}{Secs.}
\Crefname{section}{Section}{Sections}
\Crefname{table}{Table}{Tables}
\crefname{table}{Tab.}{Tabs.}

\def\cvprPaperID{4910} %
\def\confName{CVPR}
\def\confYear{2022}

\newcommand{\RNum}[1]{\uppercase\expandafter{\romannumeral #1\relax}}
\newcommand{\gpvone}{\mbox{\textsc{{GPV-1}}\xspace}}
\newcommand{\gpvtwo}{\mbox{\textsc{{GPV-2}}\xspace}}

\newcommand{\grit}{\mbox{\textsc{{GRIT}}\xspace}}

\newcommand{\coco}{\mbox{\textsc{{Coco}}\xspace}}

\definecolor{Color1}{HTML}{E6B8AF}
\definecolor{Color1b}{HTML}{FFD4CB}

\definecolor{Color2}{HTML}{F2E6C4}
\definecolor{Color2b}{HTML}{FFF2CC}

\definecolor{Color3}{HTML}{C6DAEC}
\definecolor{Color3b}{HTML}{D7EDFF}

\definecolor{Color4}{HTML}{FCCBEA}
\definecolor{Color4b}{HTML}{FFEAF5}
\definecolor{Color4c}{HTML}{F5DBE6}

\definecolor{Color5}{HTML}{F0F0F0}
\definecolor{Color5b}{HTML}{E0E0E0}

\definecolor{tabindex}{HTML}{888888}

\begin{document}

\title{GRIT: General Robust Image Task Benchmark}

\author{
Tanmay Gupta$^1$ \quad \quad 
Ryan Marten$^2$ \quad \quad 
Aniruddha Kembhavi$^1$ \quad \quad
Derek Hoiem$^2$ \quad \quad \\
$^1$PRIOR @ Allen Institute for AI \quad \quad $^2$University of Illinois at Urbana-Champaign\\
 \url{https://grit-benchmark.org}
}

\twocolumn[{%
\renewcommand\twocolumn[1][]{#1}%
\maketitle
}]

\begin{abstract}
    Computer vision models excel at making predictions when the test distribution closely resembles the training distribution. Such models have yet to match the ability of biological vision to learn from multiple sources and generalize to new data sources and tasks. To facilitate the development and evaluation of more general vision systems, we introduce the General Robust Image Task (GRIT) benchmark.  GRIT evaluates the performance, robustness, and calibration of a vision system across a variety of image prediction tasks, concepts, and data sources. The seven tasks in GRIT are selected to cover a range of visual skills: object categorization, object localization, referring expression grounding, visual question answering, segmentation, human keypoint detection, and surface normal estimation. GRIT is carefully designed to enable the evaluation of robustness under image perturbations, image source distribution shift, and concept distribution shift. By providing a unified platform for thorough assessment of skills and concepts learned by a vision model, we hope GRIT catalyzes the development of performant and robust general purpose vision systems.%
\end{abstract}

\begin{figure}[htp]
    \centering
    \includegraphics[width=\columnwidth]{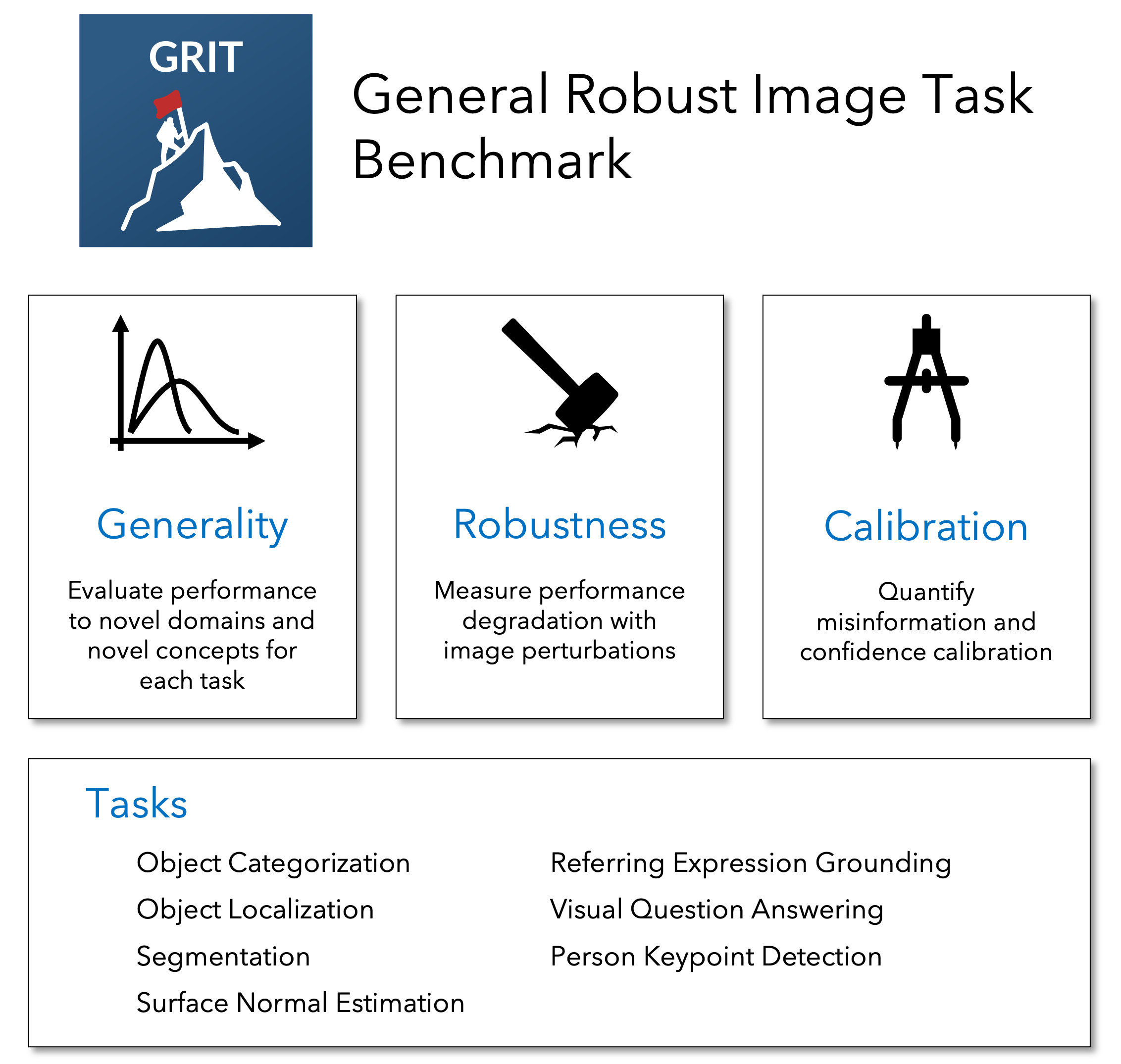}
    \caption{\textbf{A benchmark for more general vision systems.} GRIT tests the generality, robustness, and calibration of a vision system across 7 vision and vision-language tasks, multiple data sources, and diverse concepts.}
    \label{fig:grit_summary}
\end{figure}

\section{Introduction}
How do we know if a person truly understands a visual concept like ``sheep"? At the very least, we expect the person to be able to apply their conceptual understanding to a wide range of skills - identify something as a sheep, locate sheep, segment the pixels of a sheep, localize a referred-to sheep from a flock, answer simple questions about sheep, and so on. We also expect the individual to be equally capable at applying these skills on social media images, web cam footage, stock photos, and other sources. In spite of tremendous progress in computer vision, the flexibility and generality of vision systems still falls well short of this human ability. Our state-of-the-art classification systems cannot segment objects and our best segmentation models cannot answer simple questions about objects for which they are so adept at identifying pixels. Furthermore, most vision models are trained and evaluated on a limited set of concepts and under a strong i.i.d. assumption, meaning that the images and annotations in training and test sets follow the same distribution. One of the barriers to developing more flexible and general computer vision systems is the lack of a standard methodology and benchmark for evaluating performance under distribution shifts across multiple tasks and a diverse set of concepts.

The General Robust Image Task Benchmark (GRIT) evaluates the performance and robustness of a vision system across a variety of image prediction tasks, concepts, and data sources. GRIT includes seven tasks: object categorization, object localization, referring expressions, visual question answering, semantic segmentation, human keypoint estimation, and surface normal estimation. These tasks are selected to cover a range of visual skills, and evaluation includes ability to make predictions for concepts that are learned from other data sources or tasks, robustness to image perturbations, and calibration measures. 

Specifically, GRIT fulfills the following needs in development and evaluation of general robust vision systems:
\begin{itemize}
    \item GRIT is a unified platform for assessing the overall capability of computer vision systems in terms of 7 core competencies across a wide range of concepts and data sources - similar to GLUE~\cite{wang-etal-2018-glue-acl} for natural language processing systems. 
    \item GRIT tests generalization to new data sources and concepts. In contrast, most existing benchmarks only evaluate in i.i.d. settings. In the GRIT Restricted track, the model is tested on data sources and concepts that are not allowed to be seen during training. To perform tasks with novel concepts the model must transfer skills across concepts~\cite{gpv1}.
    \item GRIT tests robustness to image perturbations. Performance on each task is evaluated on a set of samples with and without 20 different types of image distortions of varying intensities such as JPEG compression, motion blur, Gaussian noise, and universal adversarial perturbation \cite{hendrycks2019robustness, dezfooli2017universal}.
    \item GRIT simultaneously supports the development of large scale foundation models for vision and the fair comparison between models with limited compute resources. This is accomplished by GRIT's two tracks: Restricted and Unrestricted. The Unrestricted track supports the development of large models with limited restrictions on allowed training data (any data source except those used to create the GRIT ablation and test sets). The Restricted track allows researchers to focus on skill-concept transfer and efficient learning given a rich but restricted set of training data sources. By limiting the training data to the selected publicly available sources, the Restricted track levels the playing field for researchers with respect to compute resource requirements and access to the same data sources. 
\end{itemize}

\section{Dataset Design Principles}
All design decisions involved in creating GRIT are motivated by the following design principles:
\begin{itemize}
\item \textbf{Unambiguous Tasks.} Following GLUE~\cite{wang-etal-2018-glue-acl}, we select vision and vision-language tasks with a clear task definition and unambiguous ground truth. We exclude captioning, for instance, as there may be many ways to caption an image. While we include the VQA task, which also has multiple possible answers, we follow the evaluation strategy used by the VQA benchmarks~\cite{Agrawal2015VQAVQ,goyal2017cvpr_balanced_vqa_v2} to eliminate ambiguity to the extent possible by including multiple answer options and answer text normalization. Additionally, we select questions with high answer consensus among annotators. 
\item \textbf{Generality and Robustness Tests}: Each task includes evaluation samples that come from different data sources, include concepts that are not present in the training data for the task, and contain image perturbations. This measures transfer ability across data sources and concepts, and robustness to image distortions. 
\item \textbf{Concept Diversity and Balance}: Task samples are selected to cover a wide and equally distributed range of concepts. Objects (noun concepts) have further been grouped into 24 concept groups (e.g. animals, food, tools).  
\item \textbf{Per-Sample Evaluation}: All metrics are computed at a sample-level so they can be averaged across various subsets of data (e.g. samples from novel sources or those containing novel concepts) to summarize performance. 
\item \textbf{Knowledge Assessment and Calibration}: Models are required to predict a confidence score for each prediction which is used to assess model's knowledge, degree of misinformation, and calibration of beliefs. 
\item \textbf{Use Existing Datasets}: When possible, we source tasks from existing, well established datasets to ensure that annotations and tasks are vetted. We also opt for hidden or even unused annotations from the selected sources. For e.g. we use COCO test-reserve annotations which are neither public nor used in any previous COCO or VQA challenges (VQA v2~\cite{goyal2017cvpr_balanced_vqa_v2} is based on COCO images). 
\item \textbf{Level Playing Field}: A Restricted track with a fixed set of publicly available training data sources allows fair comparison across submissions and enables researchers with limited access to data and compute resources to participate and contribute novel, robust, and efficient learning methods. 
\item \textbf{Encourage Unified Models}: We require all submissions to include total parameter count of the models used. While participants are allowed to use completely separate models for different tasks, we encourage models that share parameters across tasks and thus have a lower parameter count. It also serves as a simple albeit imperfect measure of compute and sample efficiency.
\end{itemize}

\section{Task Overview}
\begin{figure*}[htp]
    \centering
    \includegraphics[width=0.8\textwidth]{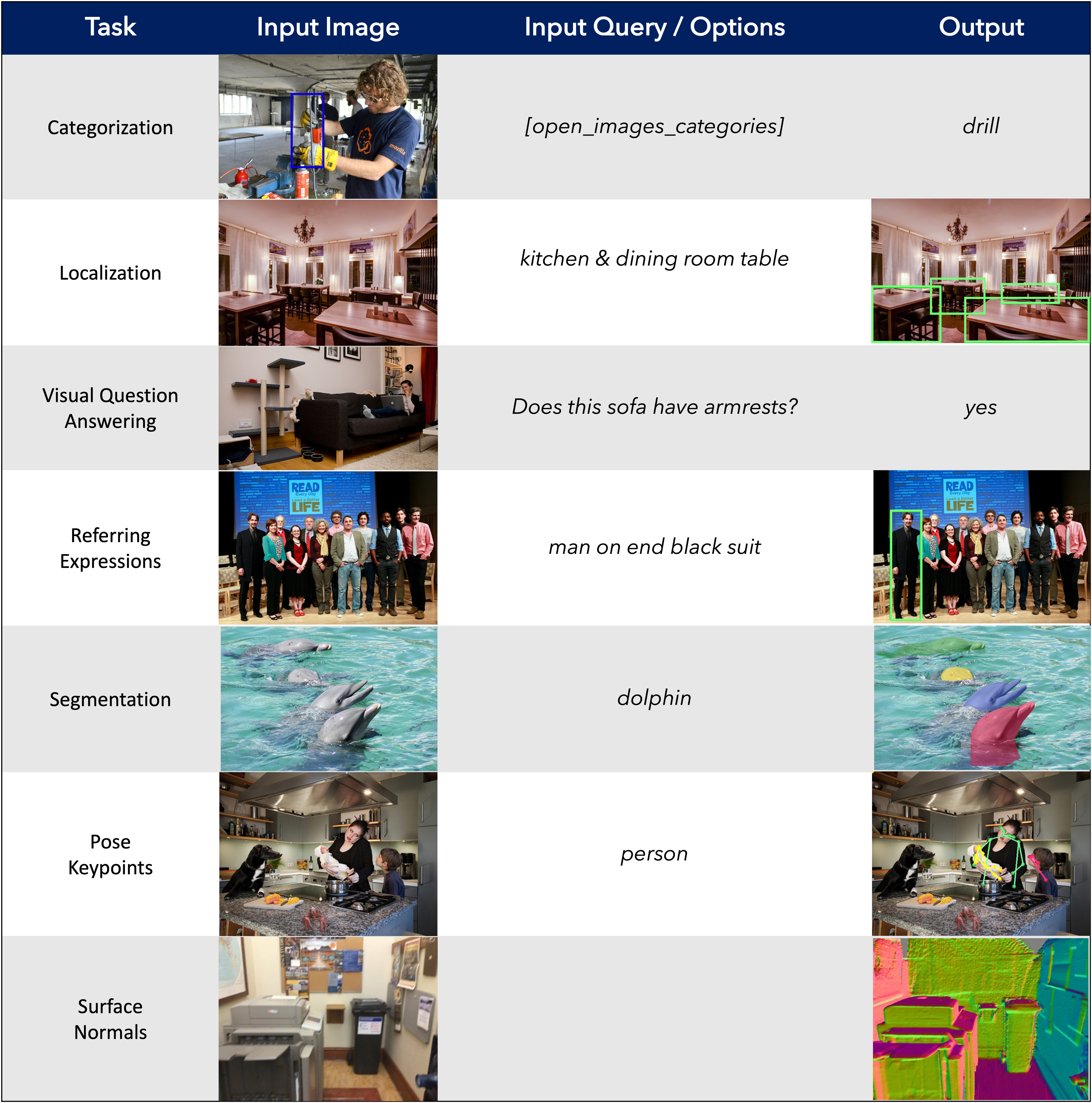}
    \caption{\textbf{Inputs and ground truth task outputs} for each of the 7 task in \grit. For the categorization task, instead of an input query, we provide a list of categories to choose from.}
    \label{fig:teaser}
\end{figure*}
\begin{table*}[t!]
\centering
\renewcommand{\arraystretch}{1.2}
\resizebox{0.8\linewidth}{!}{
\begin{tabular}{ll|c|cccc|cc}
                          &                &                 & \multicolumn{4}{c|}{\textbf{samples}}                                                & \multicolumn{2}{c}{\textbf{concepts}}   \\
\textbf{subset}           & \textbf{task}  & \textbf{images} & \textbf{total} & \textbf{novel source} & \textbf{novel concept} & \textbf{distorted} & \textbf{nouns} & \textbf{grouped nouns} \\ \hline
\multirow{7}{*}{ablation} & categorization & 12954                       & 16839 & 12478        & 10740         & 385                            & 745       & 727              \\
                          & localization   & 17457                       & 21078 & 17193        & 15112         & 385                            & 989       & 953              \\
                          & vqa            & 16017                       & 21166 & 3565         & 713           & 385                            & 9292      & 5488             \\
                          & refexp         & 4698                        & 10525 & 2781         & 935           & 385                            & 4899      & 3122             \\
                          & segmentation   & 10366                       & 12745 & 8405         & 6977          & 385                            & 695       & 680              \\
                          & keypoint       & 5385                        & 5385  & 2529         & 0             & 385                            & 1         & 1                \\
                          & normal         & 1786                        & 1786  & 675          & 0             & 385                            & 0         & 0                \\ \hline
\multirow{7}{*}{test}     & categorization & 13076                       & 16841 & 12504        & 10755         & 385                            & 766       & 752              \\
                          & localization   & 17380                       & 21080 & 17289        & 15230         & 385                            & 992       & 952              \\
                          & vqa            & 15988                       & 21166 & 3624         & 684           & 385                            & 9259      & 5441             \\
                          & refexp         & 4724                        & 10526 & 2865         & 1008          & 385                            & 5040      & 3194             \\
                          & segmentation   & 10420                       & 12730 & 8385         & 6973          & 385                            & 713       & 690              \\
                          & keypoint       & 5385                        & 5385  & 2471         & 0             & 385                            & 1         & 1                \\
                          & normal         & 1787                        & 1787  & 679          & 0             & 385                            & 0         & 0               
\end{tabular}}
\caption{\textbf{Number of images, samples, and concepts per task in \grit.}} 
\label{tab:task_stats}
\end{table*}

For each task, the system should produce both an answer (text, boxes, segmentation masks, or surface normal maps) and a confidence (score between 0 to 1) reflecting the model's belief in the correctness of the answer. A correctness score (0 to 1) is calculated by comparing the model's predicted answer to the ground truth annotation. Ideally, for a well calibrated model, the confidence would equal the correctness score. We now describe each task and the corresponding correctness score.

\textbf{Object categorization} identifies which label, from a given set, best corresponds to an image region defined by an input image and bounding box.  Different from many benchmarks, such as ImageNet~\cite{Dong09ImageNet}, the objects are typically depicted as part of a scene, and the bounding box indicates which object is of interest. A set of mutually exclusive categories representing possible answer options is provided as input to remove ambiguity. The correctness score is 1 if the prediction matches the ground truth label and 0 otherwise. 

\textbf{Object localization} places a bounding box around each instance of a given object category in the input image (or none if no objects in the target class are present). To calculate the correctness score, $P$ predicted boxes are assigned to the $G$ ground truth boxes using a Hungarian matching algorithm. Each predicted box can be assigned to a single ground truth box or not assigned at all, and at most one predicted box is assigned to each ground truth box.  The correctness score is defined as 
\begin{equation}
\sum_{i=1}^M \mathrm{IoU}_i / (P+G_{missed})    
\end{equation}  
where $\mathrm{IoU}_i$ is the intersection over union of the $i^{th}$ matched pair of ground truth and predicted boxes out of $M$ matched pairs with a non-zero IoU.  $G_{missed}$ ($=G - M$) is the number of ground truth boxes that are not assigned to any prediction.  Approximately 30\% of the images will contain zero  target class objects.  If there are no ground truth boxes, the score is 1 if there are no predictions and 0 if there are any predictions. 

\textbf{Segmentation} identifies which pixels in the input image belong to a given category. Each task specifies one class that should be segmented.  The task can be instance level (e.g. “toilet”) or stuff-level (“wall (Stuff)”). A large majority of samples are for segmenting object instances.  Instance-level segmentation expects a pixel mask for each instance in the input image and assigns these predictions to the ground truth using a Hungarian matching algorithm. Stuff-level segmentation expects a single mask for the entire image; if multiple predicted instances are provided, we use the union of these masks as the prediction during scoring. We evaluate using Boundary IoU~\cite{cheng_boundaryIoU_cvpr_2021}, which measures the intersection over union areas of the dilated boundaries of the prediction and ground truth. Boundary IoU requires more accurate segmentation of large objects without further penalizing predictions of very small objects.  The dilation parameter is $2\%$ of the image diagonal. The final correctness score, similar to localization, is 
\begin{equation}
\sum_{i=1}^M \mathrm{BoundaryIoU}_i / (P+G_{missed}).    
\end{equation}

 \begin{table*}[]
\renewcommand{\arraystretch}{1.2}
\resizebox{\linewidth}{!}{
\begin{tabular}{llc|cccc|cccc}
                                &                 & \multicolumn{1}{l|}{}                      & \multicolumn{4}{c|}{\textbf{ablation}}                                       & \multicolumn{4}{c}{\textbf{test}}                                            \\
\textbf{task}                   & \textbf{source} & \multicolumn{1}{l|}{\textbf{novel source}} & \textbf{images} & \textbf{samples} & \textbf{nouns} & \textbf{grouped nouns} & \textbf{images} & \textbf{samples} & \textbf{nouns} & \textbf{grouped nouns} \\\hline
\multirow{3}{*}{categorization} & COCO~\cite{Lin14Microsoft}          & \multicolumn{1}{l|}{} & 3804   & 4361    & 80    & 80            & 3780   & 4337    & 80    & 80             \\
                                & Open Images v6~\cite{Kuznetsova18OpenImages} & \checkmark                     & 8542   & 10014   & 475   & 469           & 8677   & 10154   & 479   & 472            \\
                                & NYU v2~\cite{silberman_2012eccv_nyuv2}         & \checkmark                     & 608    & 2464    & 351   & 338           & 619    & 2350    & 369   & 361            \\\hline
\multirow{3}{*}{localization}   & COCO~\cite{Lin14Microsoft}           & \multicolumn{1}{l|}{} & 3642   & 3885    & 80    & 80            & 3541   & 3791    & 80    & 80             \\
                                & Open Images v6~\cite{Kuznetsova18OpenImages} & \checkmark                     & 13183  & 14422   & 602   & 584           & 13208  & 14435   & 602   & 584            \\
                                & NYU v2~\cite{silberman_2012eccv_nyuv2}         & \checkmark                     & 632    & 2771    & 511   & 491           & 631    & 2854    & 519   & 495            \\\hline
\multirow{3}{*}{vqa}            & VQA v2~\cite{goyal2017cvpr_balanced_vqa_v2}         & \multicolumn{1}{l|}{} & 14061  & 17601   & 8038  & 4691          & 14051  & 17542   & 7914  & 4598           \\
                                & DAQUAR \cite{malinowski2014daquar}        & \checkmark                     & 497    & 1079    & 651   & 500           & 508    & 1127    & 672   & 499            \\
                                & DCE-VQA~\cite{Kamath2022WeblySC}        & \checkmark                     & 1459   & 2486    & 2424  & 1682          & 1429   & 2497    & 2505  & 1715           \\\hline
\multirow{3}{*}{refexp}         & RefCOCO+~\cite{kazemzadeh2014referitgame}       & \multicolumn{1}{l|}{} & 1492   & 3748    & 1904  & 1254          & 1482   & 3611    & 1912  & 1244           \\
                                & RefCOCOg~\cite{mao2016generation}       & \multicolumn{1}{l|}{} & 2211   & 3996    & 2757  & 1908          & 2233   & 4050    & 2826  & 1944           \\
                                & RefCLEF~\cite{kazemzadeh2014referitgame}        & \checkmark                     & 1080   & 2781    & 1359  & 832           & 1099   & 2865    & 1410  & 843            \\\hline
\multirow{3}{*}{segmentation}   & COCO~\cite{Lin14Microsoft}           & \multicolumn{1}{l|}{} & 4019   & 4340    & 80    & 80            & 4024   & 4345    & 80    & 80             \\
                                & NYU v2~\cite{silberman_2012eccv_nyuv2}         & \checkmark                     & 621    & 2403    & 417   & 406           & 623    & 2337    & 429   & 410            \\
                                & Open Images v6~\cite{Kuznetsova18OpenImages} & \checkmark                     & 5726   & 6002    & 328   & 323           & 5773   & 6048    & 335   & 330            \\\hline
\multirow{2}{*}{keypoint}       & COCO~\cite{Lin14Microsoft}           & \multicolumn{1}{l|}{} & 2856   & 2856    & 1     & 1             & 2914   & 2914    & 1     & 1              \\
                                & Construction~\cite{roberts2019construction, yang2016construction }   & \checkmark                     & 2529   & 2529    & 1     & 1             & 2471   & 2471    & 1     & 1              \\\hline
\multirow{4}{*}{normal}         & NYU v2~\cite{silberman_2012eccv_nyuv2}        & \checkmark                     & 331    & 331     & 0     & 0             & 323    & 323     & 0     & 0              \\
                                & BlendedMVS~\cite{yao2020blendedmvs}     & \multicolumn{1}{l|}{} & 338    & 338     & 0     & 0             & 368    & 368     & 0     & 0              \\
                                & ScanNet~\cite{dai2017scannet, huang2019framenet}        & \multicolumn{1}{l|}{} & 773    & 773     & 0     & 0             & 740    & 740     & 0     & 0              \\
                                & DTU~\cite{jensen2014large}            & \checkmark                     & 344    & 344     & 0     & 0             & 356    & 356     & 0     & 0                     
\end{tabular}}
\caption{\textbf{Number of images, samples, and concepts for each data source in \grit.}}
\label{tab:dataset_stats}
\end{table*}

\textbf{Referring expressions} places a bounding box around the instance corresponding to the provided description and image. There is always one ground truth bounding box, so only one bounding box should be predicted.  This task evaluates ability to interpret relationships and attributes received in natural language, as well as localization and categorization. The score is $1$ if $IoU > 0.5$ between the predicted box and ground truth box. Otherwise, the score is $0$. If more than one box is predicted, we use the first box in the prediction list to compute correctness and ignore the rest. 

\textbf{Visual question answering} responds with a natural language answer to an image and natural language question. Often, as for the VQA dataset~\cite{goyal2017cvpr_balanced_vqa_v2}, multiple ground truth answers are available.  If at least three answers match the prediction, the score is 1.  Otherwise, the score is the number of matches divided by three.  For example, if the annotated answers to ``What color is the dog?'' are \{brown, brown, black, brown, brown\}, then ``brown'' scores 1, ``black'' scores 1/3, and ``white'' scores 0. Questions from the original datasets were filtered to select those with at least some threshold level of ground truth annotation consistency. We use the implementation from the VQAv2~\cite{goyal2017cvpr_balanced_vqa_v2} benchmark which normalizes the answers through word contractions and removes articles and punctuation before computing the correctness score.

\textbf{Person keypoint detection} predicts pixel positions for the 17 keypoints of a human body. The correctness score is the average of keypoint scores (OKS) computed for each ground truth person instance in the image. As defined for the COCO challenge~\cite{Lin14Microsoft}, and described in the  challenge text: $\text{OKS}=\sum_i[exp(-d_i^2/2s^2\kappa_i^2)\delta(v_i>0)]/\sum_i[\delta(v_i>0)]$. 
The $d_i$ are the Euclidean distances between each corresponding ground truth and detected keypoint, and the $v_i$ are the visibility flags (0 is not labeled, 1 is labeled but not visible, 2 is labeled and visible) of the ground truth. To compute OKS, we pass the $d_i$ through an unnormalized Guassian with standard deviation $s\kappa_i$, where $s$ is the object scale and $\kappa_i$ is a per-keypoint constant that controls falloff. For each keypoint this yields a keypoint similarity that ranges between 0 and 1. These similarities are averaged over all labeled keypoints  ($v_i>0$). Predicted keypoints that are not labeled ($v_i=0$) do not affect the OKS. Perfect predictions will have $\text{OKS}=1$, and predictions for which all keypoints are off by more than a few standard deviations $s\kappa_i$ will have $\text{OKS}\approx0$. The predicted person instances are assigned to the ground truth person instances in a manner similar to the localization and segmentation tasks; OKS is computed for each combination assigning a prediction to a ground truth and the Hungarian matching algorithm selects the assignment which maximizes the total correctness score.

\textbf{Surface normal prediction} provides 3D surface normal directions for each pixel. 
Surface normal prediction and normalized depth prediction are popular geometric computer vision tasks. We choose surface normal prediction over depth regression because surface normals describe the shape of an object or surface, regardless to the distance to the camera. Additionally, we choose to have participants predict surface normals directly rather than computing them from depth because small errors in depth predictions may lead to large errors in surface normal estimates. However, we compute ground truth surface normals based on ground truth depth images, which are accurate enough to produce reasonable normals. Past works approach surface normal estimation in a scene-centric way, e.g. by predicting labels of horizontal and vertical surfaces~\cite{hoiem_surfacelayout_ijcv_2007} or major layout components~\cite{hedau_layout_iccv_2009}, or in a view-sensitive way by directly predicting the normal relative to the camera~\cite{eigen15predictdepth}. Both representations are useful depending on whether the goal is a stable representation for physical reasoning or camera-sensitive reference frame for grasping.  The correctness score accommodates both by adjusting for global orientation before computing the percent of predicted normals within 11.25 degrees of ground truth normals. The adjustment is obtained by solving for the rotation matrix that most closely aligns the predicted normals with ground truth normals. This way, normals can be predicted either from the camera's viewpoint or a scene coordinate system. %

\section{Challenge}
\begin{table}
\small
\renewcommand{\arraystretch}{1.2}
\centering
\resizebox{\linewidth}{!}{
\begin{tabular}{lll}
\toprule
\textbf{Data Source}         & \textbf{Split}                  & \textbf{Annotations}                                                                               \\ \midrule
COCO~\cite{Lin14Microsoft}                 & Train \& Val             & \makecell[l]{Bounding Boxes, Object Labels, \\ Segmentation Masks, Pose Keypoints, \\Captions}           \\ \hline
RefCOCO+~\cite{kazemzadeh2014referitgame}             & Train \& Val (UNC) & \makecell[l]{Referring Expressions, \\Bounding Boxes}                                                  \\ \hline
VQA v2~\cite{goyal2017cvpr_balanced_vqa_v2}               & Train \& Val             & Questions and Answers                                                                     \\ \hline
BlendedMVS~\cite{yao2020blendedmvs}          & Train (Our)     & Surface Normals                                                                           \\ \hline
ScanNet~\cite{dai2017scannet, huang2019framenet}             & Train                  & Surface Normals                                                                           \\ \hline
ImageNet~\cite{Dong09ImageNet}        & Any                    & Object Labels                                                                             \\ \hline
Conceptual Captions \cite{Sharma18Conceptual} & Any                    & Captions                                                                                  \\ \hline
VisualGenome~\cite{krishnavisualgenome}       & Any                    & \makecell[l]{Bounding Boxes, Object, \\Attribute, Relationship Labels \\(VQA annotations not allowed)}  \\ \hline
Web10K~\cite{Kamath2022WeblySC}              & Any                    & Queries \\ \bottomrule                                                                                  
\end{tabular}}
\caption{\textbf{GRIT Restricted training data.} To participate in the Restricted Challenge, participants \emph{must only} choose training data from one or more of the above data sources. For each source, the table specifies the splits and annotations that may be used. For BlendedMVS, we provide our own train split, where train is a subset of BlendedMVS train, and ablation/test are a subsets of BlendedMVS validation.  Other BlendedMVS train data may be used but may be less relevant (e.g. aerial photos).  Use of language models and training based on any purely non-visual data, such as Book Corpus~\cite{Zhu_2015_ICCV}, is also allowed.}
\label{tab:restricted}
\end{table}

\begin{table}
\small
\renewcommand{\arraystretch}{1.2}
\centering
\resizebox{\linewidth}{!}{
\begin{tabular}{lll}
\toprule
\textbf{Data Source}          & \textbf{Split}                     & \textbf{Annotations}                      \\ \midrule
COCO~\cite{Lin14Microsoft}                           & Test-Reserve & Any                                       \\ \hline
Open Images v6~\cite{Kuznetsova18OpenImages}                    & Test                               & Any                                       \\ \hline
NYU v2~\cite{silberman_2012eccv_nyuv2}                         & Test                               & Any                                       \\ \hline
\makecell[l]{RefCOCO+/g, RefClef}~\cite{mao2016generation,kazemzadeh2014referitgame}  & Test (UNC)                         & \makecell[l]{Referring Expressions, \\Bounding Boxes}  \\ \hline
VQA v2~\cite{goyal2017cvpr_balanced_vqa_v2}                         & Test-Reserve & Questions and Answers                     \\ \hline
DAQUAR~\cite{malinowski2014daquar}                        & Test                               & Questions and Answers                     \\ \hline
VisualGenome~\cite{krishnavisualgenome} & Test (BUTD~\cite{Anderson2018BottomUpAT})                        & Questions and Answers                                       \\ \hline
BlendedMVS~\cite{yao2020blendedmvs}                    & Test (Our)                         & Surface Normals                           \\ \hline
ScanNet~\cite{dai2017scannet, huang2019framenet}                        & Test                               & Surface Normals                           \\ \hline
DTU~\cite{jensen2014large}                           & Test                               & Surface Normals \\ \hline
Construction~\cite{roberts2019construction, yang2016construction}                  & Test (Our)                         & Pose Keypoints \\
\bottomrule                         
\end{tabular}}
\caption{\textbf{GRIT ablation and test sources.} This table lists the splits and annotations that that were used to create the GRIT ablation and test sets. Therefore, submissions to either the Restricted or the Unrestricted challenges \textit{may not} use these splits and annotations for training.}
\label{tab:unrestricted}
\end{table}

\textbf{Train Set}: The GRIT challenge consists of two tracks depending on the allowed training data. In the {\em Restricted} track, participants \emph{must only} use one or more of the data sources and annotations listed in Tab.~\ref{tab:restricted} for model tuning and hyperparameter selection. In the {\em Unrestricted} track, \textit{any} data source may be used {\em except} the sources and annotations listed in Tab.~\ref{tab:unrestricted} that were used to create the GRIT ablation and test sets. Even unsupervised learning is not allowed on the excluded sources.

\textbf{Ablation and Test Sets}: Images and task inputs are provided for the ablation and test sets. Models \emph{must not} use any direct information about the image source (e.g by identifying data source from the image ids and feeding it as an input to the model), and the ablation and test images should not be used in training or parameter selection in any way. Task-level and dataset-level statistics for GRIT ablation and test sets are shown in Tab.~\ref{tab:task_stats} and Tab.~\ref{tab:dataset_stats}.

\textbf{Leaderboards}: GRIT has an ablation and a test leaderboard per track. Participants may obtain ablation scores unlimited number of times but test scores once in a 7 day period. The test leaderboard should be used only for the final results of the main system. Submissions to both ablation and test are private until made public by the participant. Test scores are hidden unless made public.  

\section{Concepts}

To enable evaluation of core computer vision competencies across a wide range of concepts, GRIT aims to maximize the coverage of concepts while preventing over-representation of any concept. to achieve this, we first tag each sample with concepts present in the input or output text. Then, we follow a concept-based sampling strategy that for each concept identified in the source dataset, includes at least one sample containing the concept. We further cap the maximum number of samples per concept unless we encounter a sample where the concept co-occurs with another under-represented concept. A large number of concepts in GRIT are grouped into higher-level concept groups (Tab.~\ref{tab:cgroups}) that may be of interest to various application domains such as ``food", ``clothing", ``animals" etc. GRIT summarizes performance of vision systems on 24 such application domains for each task (except the keypoint task which is limited to ``people" and the surface normal task which does not have any tagged concepts).  \\

\noindent\textbf{Concept Tagging.} Each sample in \grit\ is tagged with a set of concepts (nouns, adjectives, and verbs) that appear in the task query (an object category, a VQA question, or a referring expression) or the ground truth output text (an object category or VQA answers). To tag concepts in any text, we first tokenize the text and tag each token with a POS tag. Next, we combine any consecutive noun-noun tokens (e.g. dinner table) and consecutive adj-noun tokens (e.g. hot dog) with a high ($> 0.35$) normalized pointwise mutual information (NPMI) into compound nouns. We use the unigram and bigram frequencies from BERT's~\cite{Devlin19Bert} training corpus (BookCorpus~\cite{zhu_bookcorpus_2015iccv} and Wikipedia) to compute NPMI. All nouns, compound nouns, adjectives, and verbs are collected as concept tags. Each tag consists of the original and lemmatized text as well as one of 4 tags - NOUN, CNOUN, ADJ, and VERB. \\  

\noindent\textbf{Concept Grouping.} A large number of noun and compound nouns in \grit\ are grouped into at least one of 24 concept groups (Tab.~\ref{tab:cgroups}). We used Amazon Mechanical Turk (AMT) to map lemmas to concept groups. Specifically, for any lemma $\mathcal{L}$ that appears more than once in \grit\, we ask 3 workers whether ``$\mathcal{L}$ belongs to the group of $\mathcal{C}$" for every concept group $\mathcal{C}$. We then compute an agreement score among the 3 workers for the hypothesis $\mathcal{L} \in \mathcal{C}$ as the sum of worker-quality weighted binary assignment score. The worker-quality is computed as the fraction of annotations where the worker's answer matched the majority-vote answer. The assignment is accepted if the agreement score exceeds a chosen threshold. Lemmas that appear only once in \grit\ and whose head nouns have already been annotated through AMT, borrow the head noun's assignments. Remaining single-occurrence lemmas are annotated via AMT. \\

\begin{table*}[]
\resizebox{\linewidth}{!}{
\begin{tabular}{rrl}
\toprule
\textbf{concept group} & \textbf{\#concepts} & \textbf{concept lemmas (sampled)}                                                                      \\ \midrule
food                                       & 1278                                    & artichoke, lamb, powdered sugar, cake, common fig, banana, hot dog, pizza, peach, rice                     \\
people                                     & 1056                                    & man, person, driver, skateboarder, guy, race, boy, sweater guy, snowboarder, cook                          \\
places                                     & 1005                                    & bar, side, computer room, building, basement, convenience store, lighthouse, hill, pet store, porch        \\
kitchen\_objects                           & 961                                     & sink, spoon, bottle, bowl, platter, wine bottle, fridge, stove, water bottle, cabinet                      \\
animals                                    & 847                                     & alpaca, food, animal, cat, dog, clydesdale, fox, dragonfly, bee, beak                                      \\
clothing                                   & 756                                     & color clothe, tie, jacket, shoe, shirt, jean, top, maroon shirt, black tshirt, coat                        \\
structure                                  & 703                                     & room, wall, fence, socket, fountain, road, awning, sign, pool, construction area                           \\
vehicles                                   & 680                                     & plane, bus, airplane, motorcycle, bike, bicycle, old car, transportation, barge, submarine                 \\
household\_objects                         & 668                                     & clock, luggage, picture, pillow, scissor, vase, photo, telephone, bottom plant, light bulb                 \\
technology                                 & 522                                     & turbine, iphone, water purifier, information, mobile phone, microwave, remote, mouse, TV screen, apple     \\
sports\_equipment                          & 514                                     & kite, bicycle, football, bicycle wheel, new jersey, frisbee, volleyball, cricket ball, tennis racket, oar  \\
body\_parts                                & 493                                     & boys foot, beard, nose, jersey head, hand, chest, head, ear, tusk, eye                                     \\
clothing\_accessories                      & 481                                     & backpack, bandana, necktie, sneaker, purse, hat, handbag, glove, high heel, mask                           \\
furniture                                  & 468                                     & metal chair, chair, bed, tile counter, table, wardrobe, couch, counter, stool, table square                \\
natural\_landscape                         & 439                                     & sand, grass, water, apple, land, wood, river, cairngorm, tree, maple                                       \\
transport\_infrastructure                  & 366                                     & left, fire hydrant, parking meter, stop sign, road, mosco street, traffic light, dock, bus route, ski lift \\
bathroom\_objects                          & 359                                     & toilet, toothbrush, floor mat, lipstick, restroom, cabinet, plastic object, towel rod, shower, sponge      \\
brands                                     & 332                                     & brand name, ford, ipad, ipod, dell, sign advertising, ducati, camel, adida, tarmak                         \\
tools                                      & 295                                     & rope, white utensil, controller, rifle, scale, object, box, drill, personal flotation, cart                \\
plants                                     & 254                                     & grass, hay, tree bush, red flower, flower, ornamental plant, straw, floret, stick, tree                    \\
stationery                                 & 251                                     & folder, letter, office supply, object, paper, ruler, tag, whiteboard, paper holder, book                   \\
beverages                                  & 206                                     & juice, water, milk, coffee, liquid, drink, healthy meal, beer, cocktail, wine                              \\
birds                                      & 140                                     & bat, bird, hummingbird, raven, goose, sparrow, egg, swan, chicken, penguin                                 \\
musical\_instruments                       & 75                                      & wind, keyboard, guitar, organ, piano, flute, violin, scale, saxophone, horn \\ \bottomrule                              
\end{tabular}}
\caption{\textbf{Concept groups} with number of unique concepts and 10 random concepts from each group sampled in proportion to the frequency of occurrence in \grit. While generally accurate, a few errors in identifying concepts and concept lemma to group mapping stem from inaccurate POS tagging, incorrect assignment by AMT workers (e.g. ``left" is mapped to ``transport\_infrastructure"), or the automatic mapping of single-occurrence lemmas using head nouns (e.g. ``jersey head", a type of head wrap, is mapped to ``body\_parts" instead of ``clothing").}\label{tab:cgroups}
\end{table*}

\begin{figure*}[t]
    \centering
    \includegraphics[trim=0 100 200 100,width=0.8\linewidth]{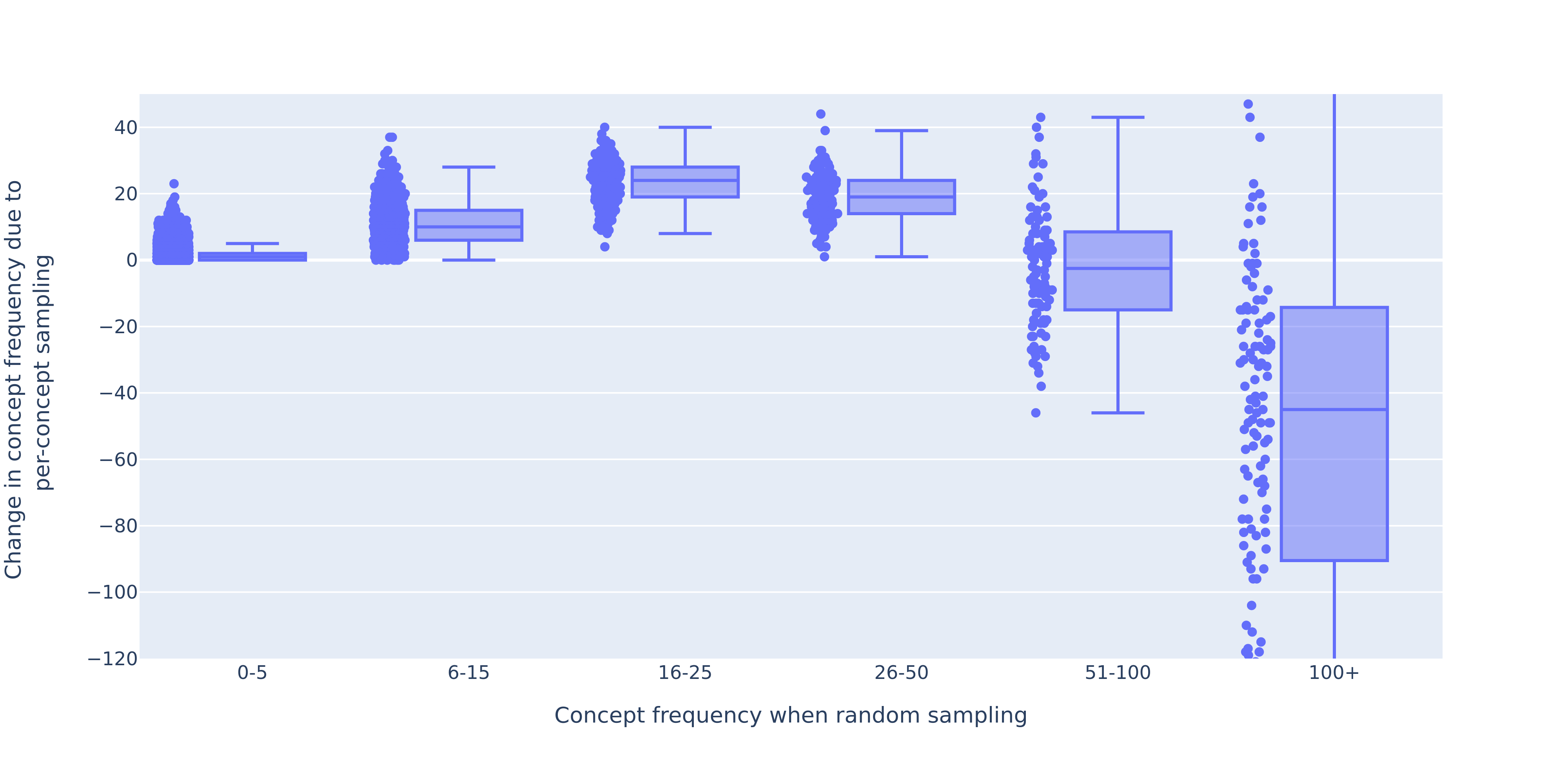}
    \caption{\textbf{Effect of per-concept sampling on VQAv2 dataset.} In the plot, each point corresponds to a concept. With $N=50$, per-concept sampling generally selects more samples for concepts that would have been represented less than 50 times under random sampling while reduces the representation of concepts that appear more than 50 times during random sampling.}
    \label{fig:sampling}
\end{figure*}

\noindent\textbf{Sampling.} Instead of sampling uniformly at random from each data source, we follow a per-concept sampling strategy that maximizes the number of concepts represented in \grit\ while also preventing any concept from dominating the evaluation. Specifically, we iterate over the noun concepts in increasing order of frequency in the original dataset and sample a fixed number of samples, say N, for each concept without replacement. Due to co-occurrence of concepts in samples,  $S_i$ samples may already have been sampled for the $i^{th}$ concept while sampling other less frequent concepts. Therefore, we only sample an additional min$(R_i,N-S_i)$ samples from the remaining $R_i$ samples for the concept. The number of remaining samples and the number of selected samples for each concept are updated every iteration after selecting the concept samples. Fig.~\ref{fig:sampling} contrasts per-concept sampling with random sampling when selecting \grit\ samples from VQAv2. The overall concept frequency distribution in the GRIT ablation set resulting from the per-concept sampling strategy is shown in Tab.~\ref{tab:concept_freq}.

\begin{table}[!h]
\centering
\resizebox{\columnwidth}{!}{
\begin{tabular}{lrrrrrrr}
\toprule
          \textbf{task} &  \textbf{0-5} &  \textbf{6-15} &  \textbf{16-25} &  \textbf{26-50} &  \textbf{51-100} &  \textbf{101-500} &  \textbf{$>$500} \\
\midrule
categorization &  204 &    84 &    213 &    170 &      69 &        5 &     0 \\
  localization &  316 &    39 &    304 &    252 &      74 &        4 &     0 \\
           vqa & 7884 &   729 &    228 &    261 &     114 &       71 &     5 \\
        refexp & 4402 &   260 &     84 &     75 &      57 &       20 &     1 \\
  segmentation &  283 &    91 &    172 &     80 &      69 &        0 &     0 \\
      keypoint &    0 &     0 &      0 &      0 &       0 &        0 &     1 \\
        normal &    0 &     0 &      0 &      0 &       0 &        0 &     0 \\
\bottomrule
\end{tabular}
}
\caption{\textbf{Concept frequency distribution.} Each column labeled $x$-$y$ indicates the number of NOUN or CNOUN concepts that appear at least $x$ and at most $y$ times for different task in the GRIT ablation set.}
\label{tab:concept_freq}
\end{table}

\section{Metrics}

Metrics for \grit\ comprise of accuracy, knowledge, and calibration measures computed on various subsets of data. Specifically, we follow a 4 part nomenclature for metrics - \textit{measure.cgroup.partition.task}. 
\begin{itemize}
    \item \textit{measure}: one of the performance measures listed in Tab.~\ref{tab:measures} and described in Sec.~\ref{sec:knowcal_measures}.
    \item \textit{cgroup}: specifies the concept group to measure performance over. \grit\ consists of 24 concept groups (e.g. ``plants", ``animals", ``tools", ``furniture" etc.) and \textit{cgroup}$=$``any" implies no concept group restriction. 
    \item \textit{partition}: restricts metric computation to samples with a specific property (e.g. samples containing distorted images). \grit\ consists of 7 \textit{partitions} described in Tab.~\ref{tab:partitions}.
    \item \textit{task}: one of the 7 tasks or ``all" for overall performance across all tasks. 
\end{itemize}

\subsection{Robustness}
\grit\ allows evaluation of robustness under distribution shifts due to the following factors:
\begin{itemize}
  \item \textbf{Change in image source}: performance may be compared across samples with the same or different image source than training data by setting: \\
  \textit{partition}$=$\{\textit{sameSrc}, \textit{newSrc}\}
  
  \item \textbf{Change in concept distribution}: performance may be compared across samples that only contain concepts seen during training or ones containing at least one novel concept by setting: \\
  \textit{partition}$=$\{\textit{sameCpt}, \textit{newCpt}\}
  
  \item \textbf{Image perturbation}: a range of image perturbations of various types and intensities have been applied to a subset of 385 samples from each task. Specifically, we apply each of the 19 types of distortion from~\cite{hendrycks2019robustness} at five severity levels to 3 samples per task, and apply universal adversarial perturbations from~\cite{Moosavi-Dezfooli17} at five severity levels to 20 samples per task. Performance may be compared on this subset for samples with and without distortions by setting: \\ \textit{partition}$=\{$\textit{dist}, \textit{undist}\}
\end{itemize}

\begin{table*}[t]
\centering
\small
\renewcommand{\arraystretch}{1.2}
\begin{tabular}{lllc}
\toprule
\textbf{Measure} & \textbf{Abbrev.} & \textbf{Purpose}            & \textbf{Function} \\ \midrule
Accuracy*         & acc                   & Correctness                 & $\sum_i a_i/N$                       \\ 
Information      & inf                   & Confident correctness         & $\sum_i a_i \cdot c_i/N$                      \\ 
Misinformation   & misinf                & Confident incorrectness       & $\sum_i c_i\cdot(1-a_i)/N$                  \\ 
Confidence       & conf                  & Actionability & $\sum_i c_i/N$                        \\ 
Self-Awareness   & sa                    & Calibration                 & $\sum_i \{c_i\cdot a_i + (1-c_i)\cdot (1-a_i)\}/N$        \\ 
RMSE*             & rmse                  & Calibration                 & $\sqrt{\sum_i (c_i-a_i)^2/N}$    \\ \bottomrule            
\end{tabular}
\caption{\textbf{Knowledge and calibration measures} computed over a set of $N$ samples indexed by $i$ with $a_i$ and $c_i$ as the accuracy (correctness score) and confidence of the prediction for the $i^{th}$ sample. Confidence and correctness score are assumed to be in the range of 0 to 1. Tab.~\ref{tab:partitions} defines the sets over which the measures are computed. * highlights the main measures to be used for model comparison on \grit.}
\label{tab:measures}
\end{table*}
\begin{table*}[!h]
\small
\centering
\renewcommand{\arraystretch}{1.2}
\begin{tabular}{lll}
\toprule
\textbf{Partition} & \textbf{Abbrev.} & \textbf{Description}                                                                            \\ \midrule
Same source        & sameSrc          & Samples that share the same image source as primary training data                               \\
New source*         & newSrc           & Samples that use an image source different from primary training data                           \\
Aggregate*          & agg              & Average of sameSrc and newSrc performance                                                       \\
Same concept       & sameCpt          & Samples that only contain concepts seen in primary training data                                \\
New concept        & newCpt           & Samples containing at least one concept unseen in primary training data                         \\
Distorted          & dist             & Samples containing distorted or perturbed images                                                \\
Undistorted        & undist           & Samples containing corresponding undistorted images                                             \\
Delta distorted    & deldist          & Distorted and undistorted sample pairs to compute change in measures due to distortion \\ \bottomrule
\end{tabular}
\caption{\textbf{Partitions} define various subsets of samples over which measures shown in Tab.~\ref{tab:measures} are aggregated. * indicates the main partitions to use for model comparison for \grit.}
\label{tab:partitions}
\end{table*}

\subsection{Knowledge and Calibration}\label{sec:knowcal_measures}
Typically, vision and vision-language benchmarks encourage the research community to build more accurate models. However, models are known to be confidently wrong~\cite{Li2020ImprovingCE} with potentially undesirable consequences. Further to evaluate knowledge, it is important to consider the model's certainty in its beliefs in addition to whether those beliefs are correct~\cite{Hunt2003TheCO}. To encourage models that are both accurate and well-calibrated, in addition to accuracy, \grit\ includes the following measures of model knowledge and calibration:
\begin{itemize}
    \item \textbf{Confidence}: AI-driven human decision making often hinges on the model's reported confidence in its own predictions. A low confidence prediction may be ignored by the decision maker as unreliable while a high confidence prediction is likely to be trusted and \emph{used} in the decision making process. We report average confidence as a measure of a model's \emph{usable} or \emph{actionable} beliefs. These are beliefs that may be used in the decision making process irrespective of whether those beliefs are correct.
    \item \textbf{Information}: confidence weighted correctness score that indicates how often the model is both confident and correct and hence well-informed.
    \item \textbf{Misinformation}: confidence weighted complement of correctness score that indicates how often the model is confidently wrong or misinformed. This reflects the portion of model's beliefs with potentially negative and harmful consequences with regards to fairness and operational risk (e.g in autonomous driving applications).  
    \item \textbf{Self-Awareness}: measure of whether a model knows what it knows and doesn't know.
    \item \textbf{RMSE}: root mean square error between confidence and correctness score that indicates whether the predicted confidence is a reliable proxy for accuracy.
\end{itemize}

\subsection{Recommended metrics}
While \grit\ provides a number of informative metrics for model performance analysis, for consistency, we recommend the following metrics for comparing models on individual tasks on \grit:
\begin{itemize}
    \item \{acc, rmse\}.any.\{agg, newSrc\}.\{task\}: accuracy and RMSE computed on the ``Aggregate" and ``New source" partitions for each task 
    \item acc.any.dist.\{task\}: accuracy on the distorted samples for each task
\end{itemize}

For comparing models on overall performance across all tasks, we recommend:
\begin{itemize}
    \item acc.any.agg.all: average accuracy over all combinations of \{task\}$\times$\{sameSrc,newSrc\} as the primary metric
    \item acc.any.dist.all: average accuracy on distorted samples across tasks
\end{itemize}
Note that these overall measures assume $0$ accuracy on tasks not supported by the model.

\section{Baselines}

\begin{table*}[t]
\centering
\resizebox{\linewidth}{!}{
\begin{tabular}{l|c|cc|cc|cc|cc|cc|cc|cc} 
                       &  & \multicolumn{2}{c|}{\textbf{cat}} & \multicolumn{2}{c|}{\textbf{loc}} & \multicolumn{2}{c|}{\textbf{vqa}} & \multicolumn{2}{c|}{\textbf{ref}} & \multicolumn{2}{c|}{\textbf{seg}} & \multicolumn{2}{c|}{\textbf{kp}} & \multicolumn{2}{c}{\textbf{sn}}  \\
                                \textbf{Model}  & \textbf{restricted}  & \textbf{agg} & \textbf{newSrc}    & \textbf{agg} & \textbf{newSrc}    & \textbf{agg} & \textbf{newSrc}    & \textbf{agg} & \textbf{newSrc}    & \textbf{agg} & \textbf{newSrc}    & \textbf{agg} & \textbf{newSrc}   & \textbf{agg} & \textbf{newSrc}   \\ 
\hline
Bae et al.~\cite{Bae2021EstimatingAE}      &    \checkmark       & -            & -                  & -            & -                  & -            & -                  & -            & -                  & -            & -                  & -            & -                 & 49.4        & 43.9             \\
Mask-RCNN~\cite{He17Mask} & \checkmark & -            & -             & 44.6                 & 43.3      & -                       & -                  & -            & -                  & 26.2        & 8.2               & 70.8        & 72.1              & -            & -                 \\
GPV-1~\cite{gpv1}        &    \checkmark                    & 33.2        & 9.4                & 42.8        & 39.5              & 50.6        & 38.0              & 25.8        & 18.4              & -            & -                  & -            & -                 & -            & -                 \\ \hline
GPV-2~\cite{Kamath2022WeblySC}   &                             & 54.8        & 22.9               & 53.5        & 54.7              & 63.5        & 53.3              & 51.5        & 39.4              & -        & -              & -        & -             & -        & -             \\
\end{tabular}}
\caption{\textbf{Aggregate performance and generalization to new sources on GRIT ablation set.} Same / new source partitions are true to their name only in the Restricted setting. \textbf{agg}: acc.any.agg.\textit{task}; \textbf{newSrc}: acc.any.newSrc.\textit{task}}
\label{tab:src_generality}
\end{table*}

\begin{table*}[th!]
\centering
\resizebox{\linewidth}{!}{
\begin{tabular}{l|c|cc|cc|cc|cc|cc|cc|cc} 
                       &  & \multicolumn{2}{c|}{\textbf{cat}}  & \multicolumn{2}{c|}{\textbf{loc}}  & \multicolumn{2}{c|}{\textbf{vqa}}  & \multicolumn{2}{c|}{\textbf{ref}}  & \multicolumn{2}{c|}{\textbf{seg}}  & \multicolumn{2}{c|}{\textbf{kp}}   & \multicolumn{2}{c}{\textbf{sn}}     \\
    \textbf{Model}                                 &         \textbf{restricted}            & \textbf{same} & \textbf{new} & \textbf{same} & \textbf{new} & \textbf{same} & \textbf{new} & \textbf{same} & \textbf{new} & \textbf{same} & \textbf{new} & \textbf{same} & \textbf{new} & \textbf{same} & \textbf{new}  \\ 
\hline
Bae et al.~\cite{Bae2021EstimatingAE}                 & \checkmark                & -                & -               & -                & -               & -                & -               & -                & -               & -                & -               & -                & -               & 49.6            & -                \\
Mask-RCNN~\cite{He17Mask} & \checkmark                & -                & -               & 51.9                & 40.8               & -                & -               & -                & -               & 44.9            & 0.3            & 70.9            & -               & -                & -                \\
GPV-1~\cite{gpv1}                                & \checkmark                & 58.7            & 0.8            & 48.3            & 37.8           & 58.3            & 74.0           & 29.7            & 23.1           & -                & -               & -                & -               & -                & -                \\ \hline
GPV-2~\cite{Kamath2022WeblySC}                                &                & 84.9            & 13.5           & 54.6            & 54.2           & 69.8            & 81.7           & 57.8            & 48.3            & -            & -           & -            & -           & -            & -            \\
\end{tabular}}
\caption{\textbf{Generalization to new concepts on GRIT ablation set.} Same / new concept partitions are true to their name only in the Restricted setting. \textbf{same}: acc.any.sameCpt.\textit{task}; \textbf{new}: acc.any.newCpt.\textit{task}}
\label{tab:cpt_generality}
\end{table*}

\begin{table*}[h!]
\centering
\resizebox{\linewidth}{!}{
\begin{tabular}{l|c|cc|cc|cc|cc|cc|cc|cc} 
                        &                              & \multicolumn{2}{c|}{\textbf{cat}} & \multicolumn{2}{c|}{\textbf{loc}} & \multicolumn{2}{c|}{\textbf{vqa}} & \multicolumn{2}{c|}{\textbf{ref}} & \multicolumn{2}{c|}{\textbf{seg}} & \multicolumn{2}{c|}{\textbf{kp}} & \multicolumn{2}{c}{\textbf{sn}}  \\
\textbf{\textbf{Model}} & \textbf{\textbf{restricted}} & \textbf{undist} & \textbf{dist}   & \textbf{undist} & \textbf{dist}   & \textbf{undist} & \textbf{dist}   & \textbf{undist} & \textbf{dist}   & \textbf{undist} & \textbf{dist}   & \textbf{undist} & \textbf{dist}  & \textbf{undist} & \textbf{dist}  \\ \hline
Bae et al.~\cite{Bae2021EstimatingAE}              & \checkmark                         & -               & -               & -               & -               & -               & -               & -               & -               & -               & -               & -               & -              & 54.3           & 42.6          \\
Mask-RCNN~\cite{He17Mask}               & \checkmark                         & -               & -               &  47.4              & 20.6               & -               & -               & -               & -               & 40.6            & 18.8           & 67.9            & 43.4          & -               & -              \\
GPV-1~\cite{gpv1}                  & \checkmark                         & 58.4           & 29.1           & 45.6           & 32.2           & 65.2           & 57.1           & 30.1           & 33.0           & -               & -               & -               & -              & -               & -              \\ \hline
GPV-2~\cite{Kamath2022WeblySC}                   &                         & 86.5           & 64.7           & 51.7            & 34.9           & 72.4           & 61.8           & 60.5           & 56.4           & -           & -           & -           & -          & -           & -          \\
\end{tabular}}
\caption{\textbf{Robustness to image distortions on GRIT ablation set.} For each task, performance is computed on the same set of images and model inputs with and without distortion. \textbf{undist}: acc.any.undist.\textit{task}; \textbf{dist}: acc.any.dist.\textit{task}}
\label{tab:robustness}
\end{table*}

\begin{table*}[!h]
\centering
\resizebox{\linewidth}{!}{
\begin{tabular}{l|c|cc|cc|cc|cc|cc|cc|cc} 
                                     &                     & \multicolumn{2}{c|}{\textbf{cat}} & \multicolumn{2}{c|}{\textbf{loc}} & \multicolumn{2}{c|}{\textbf{vqa}} & \multicolumn{2}{c|}{\textbf{ref}} & \multicolumn{2}{c|}{\textbf{seg}} & \multicolumn{2}{c|}{\textbf{kp}} & \multicolumn{2}{c}{\textbf{sn}}  \\
\textbf{Model}                       & \textbf{restricted} & \textbf{rmse} & \textbf{sa}       & \textbf{rmse} & \textbf{sa}       & \textbf{rmse} & \textbf{sa}       & \textbf{rmse} & \textbf{sa}       & \textbf{rmse} & \textbf{sa}       & \textbf{rmse} & \textbf{sa}      & \textbf{rmse} & \textbf{sa}      \\ 
\hline
Bae et al.~\cite{Bae2021EstimatingAE}                 & \checkmark                & -             & -                 & -             & -                 & -             & -                 & -             & -                 & -             & -                 & -             & -                & 55.5         & 49.4            \\
Mask-RCNN~\cite{He17Mask} & \checkmark                & -             & -                 & 49.1             & 66.8                 & -             & -                 & -             & -                 & 21.1          & 85.5             & 32.8         & 72.8            & -             & -                \\
GPV-1~\cite{gpv1}                                & \checkmark                & 58.0         & 49.5             & 50.1         & 51.5             & 49.0         & 60.8             & 65.7         & 44.1             & -             & -                 & -             & -                & -             & -                \\ \hline
GPV-2~\cite{Kamath2022WeblySC}&                & 55.6         & 62.5             & 42.4         & 55.1             & 37.7          & 71.0             & 48.6          & 59.4              & -         & -             & -         & -            & -         & -            \\
\end{tabular}}
\caption{\textbf{Calibration on GRIT ablation set.} \textbf{rmse}: rmse.any.agg.\textit{task} (lower is better); \textbf{sa}: sa.any.agg.\textit{task}}
\label{tab:calibration}
\end{table*}

We present a preliminary set of experiments to demonstrate various evaluation capabilities afforded by GRIT and to highlight avenues for future research. Since there is no single model that can perform all the GRIT tasks, we evaluate the following models to cover all tasks. 
\begin{enumerate}
    \item \textbf{\gpvone}~\cite{gpv1} is a task-agnostic vision language architecture that can perform any task with text and image as inputs, and text, bounding boxes and region relevance scores as outputs. \gpvone\ is trained on \coco\ for categorization, localization, VQA, and captioning tasks. We evaluate \gpvone\ on the GRIT categorization, localization, and VQA tasks for which the model was trained using \coco\ (along with \coco\ captioning). We also evaluate \gpvone\ on the referring expression task in a \textit{zero-shot} setting. Confidence scores for categorization and vqa are the likelihood scores of the output text. For localization, only boxes with relevance greater than $0.5$ are selected as predictions with the average relevance being the prediction confidence. For referring expression, the most relevance box and its relevance score are used as prediction and confidence respectively.  
    
    \item \textbf{\gpvtwo}~\cite{Kamath2022WeblySC} is a GPV architecture based on the T5~\cite{Raffel2020t5} encoder-decoder architecture trained on multiple NLP tasks and uses region proposals and visual representations computed by the powerful VinVL~\cite{Zhang2021VinVLMV} object detector. \gpvtwo\ is trained to perform the same set of tasks as \gpvone\ along with a classification-in-context task and the referring expressions task but for more than 10,000 concepts instead of just the 80 primary \coco\ categories. To do so, \gpvtwo\ learns these concepts from web image search results for noun, adj-noun, and noun-verb queries and transfers the learned concepts across skills learned from task-specific \coco\ annotations. Confidence scores for \gpvtwo\ are generated in a manner similar to \gpvone, but \gpvtwo\ reuses the language decoder to score boxes.
    
    \item \textbf{Mask-RCNN}~\cite{He17Mask} is a well-tuned detection, segmentation, and keypoint prediction architecture that uses an anchor-box based region proposal network, and box, class, and segmentation heads to produce region-level outputs. Since. Mask-RCNN is limited to \coco\ categories, predicted instances in segmentation and localization are selected only if their label matches the query class. The confidence score is calculated as an average of selected instance confidence scores.  
    
    \item \textbf{Bae et al.}~\cite{Bae2021EstimatingAE} uses the estimated aleatoric uncertainty in surface normal prediction to guide the stochastic selection of pixels to use for training. Input images are resized , and the predicted normal map is resized back to the original input size and multiplied by -1 to match the ground truth coordinate frame. The confidence score is calcualted as the percentage of pixels in the uncertainty map with expected angular error less than $22.5^{\circ}$. We use the publicly available model trained on ScanNet.
    
\end{enumerate}

Except for \gpvtwo, all of the above models are eligible for submission to the Restricted track. \gpvtwo, however, can only be submitted to the Unrestricted track since it uses a VinVL backbone trained on Open Images v6 and Objects365 (along with \coco\ and Visual Genome) which are not part of the Restricted training set. We now discuss generality, robustness, and calibration of these models on the GRIT benchmark. \\

\noindent\textbf{Generality.} Tab.~\ref{tab:src_generality} shows accuracy of each model across supported tasks. Generally, performance drops on new sources across tasks for the Restricted models. On the keypoints task, model performance is higher on the new source since Construction dataset images are slightly biased towards simpler images focused on a single, clearly visible person. \gpvtwo\ shows a similar drop in new source accuracy with the exception of localization. This is because in the Unrestricted track, models are allowed to train on the Open Images train set which is one of the novel sources. Note that ``same" and ``novel" are defined with respect to the Restricted training data and may not be applicable to the evaluation of generalization to novel sources in the Unrestricted setting. 

Tab.~\ref{tab:cpt_generality} shows that Restricted models struggle to generalize to novel tasks, especially categorization and segmentation. Both of these tasks require predicting a category label never seen in the task's training data. On VQA, the performance on the new concepts is significantly higher. However, note that VQA new concept evaluation is somewhat limited since VQAv2 training data covers a surprisingly large number of concepts resulting in only a small number of samples with novel concepts many of which are of the relatively high scoring Yes/No question type (see performance breakdown by question types in Tab.2 and 3 in ~\cite{goyal2017cvpr_balanced_vqa_v2}). Finally, the VQA novel concepts are often bigrams (e.g ``winter glove", ``restaurant pizza") where one or both of the words is either superfluous or seen during training. Keypoint task is limited to a single concept (``person") and surface normal samples are not tagged with concepts, and hence these tasks do not report performance on the new concept partition. \\

\noindent\textbf{Robustness.} Tab.~\ref{tab:robustness} shows that all models show a drop in performance across all tasks due to image distortion with the exception of zero-shot \gpvone\ on referring expressions task. Note that in this case, the performance on the undistorted images is already quite low as expected from a zero-shot model. \\

\noindent\textbf{Calibration.} Tab.~\ref{tab:calibration} shows RMSE and self-awareness for each model and the supported tasks. RMSE measures how reliable is the model predicted confidence score as an estimate of model correctness. Self-awareness also encourages greater confidence scores for correct predictions and lower confidence for incorrect ones. A model may maximize self-awareness by predicting confidence $1.0$ when the correctness score is greater than $0.5$ and $0$ otherwise. However, this reduces the model's RMSE. Therefore, the choice of calibration metric to optimize depends on how you plan to use the confidence scores in downstream decision making. \\

\begin{table}[t!]
\centering
\resizebox{\columnwidth}{!}{
\begin{tabular}{l|c|c|c|c}
               \textbf{Model} &  \textbf{restricted} &  \textbf{params (M)} &  \textbf{acc.any.agg.all} &  \textbf{acc.any.dist.all} \\
\hline
Mask-RCNN~\cite{He17Mask} & \checkmark &      58 &  20.2 &   11.8 \\
Bae et al.~\cite{Bae2021EstimatingAE} &        \checkmark &      72 &  7.1 &   6.1 \\
               GPV-1~\cite{gpv1} &        \checkmark &     236 & 21.8 &  21.6 \\\hline
               GPV-2~\cite{Kamath2022WeblySC} &        &     370 & 31.9 &  31.1 \\
\end{tabular}}
\caption{\textbf{Model size and overall performance averaged across all tasks on the GRIT ablation set.} Accuracy is assumed to be $0$ for tasks not supported by the model.}
\label{tab:overall}
\end{table}
\noindent\textbf{Overall performance.} Model parameter counts and overall performance for all baselines are shown in Tab.~\ref{tab:overall}

\section{Related Generalization Benchmarks}

The \textbf{Robust Vision Challenge} \cite{RVC2020} invites researchers to submit their model's results to one or more of seven tasks (stereo, optical flow, single-view depth estimation, object detection, semantic/instance/panoptic segmentation) across 13 benchmark datasets . For any submitted task, the same model must be used to generate results on all applicable benchmarks. This tests a system's ability to learn and perform on multiple data distributions simultaneously, while GRIT, in the Restricted track, further requires ability to generalize to new data and label distributions by prohibiting training on held out benchmarks.  GRIT also includes language tasks (VQA, referring expressions) and excludes video and multiview tasks.

\textbf{ObjectNet} \cite{barbu_2019_neurips_objectnet} is a test-only dataset that evaluates generalization of object classification to a new data source and broad pose variation. GRIT offers a broader set of tasks and stipulates a standard set of training data to enable comparison of in-distribution of new-distribution performance while controlling (optionally) for training data.

\textbf{GLUE} \cite{wang-etal-2018-glue-acl} (General Language Understanding Evaluation) provides a broad benchmark to test state-of-the-art approaches on natural language processing tasks. \textbf{SuperGLUE} \cite{wang2019superglue} is a improved revision of GLUE that includes a set of more challenging language tasks. We adopt some design principles from GLUE, such as restricting to tasks with unambiguous ground truth and using existing datasets where possible. For pragmatic reasons, we do flex on these ideals, as for example questions in VQA can often legitimately be answered in multiple ways, and we create new benchmarks for person keypoint prediction and surface normal estimation to increase diversity of data sources.

\section{Conclusion}
In empirical computer vision and language understanding research, high quality benchmarks such as ImageNet, MSCOCO, and GLUE have driven progress over multiple years or even an entire decade. As a result, we now have highly performant, data-driven, task-specific models. The time is ripe to unify these advances into more general purpose systems that are flexible enough to perform a wide range of tasks without requiring architecture changes and robust enough to withstand the drafts of distribution shift that plague vision and vision-language models in the open world setting. GRIT hopes to drive the development of such models by providing a unified platform for evaluation of generality and robustness of 7 core capabilities of computer vision across multiple data sources and diverse concepts.

\section{Acknowledgements}
We thank Tsung-Yi Lin and the COCO team, Aishwarya Agrawal and the VQA team, Angela Dai and the Scannet team, and the Construction Keypoints team for valuable data contributions to GRIT. We are also thankful to Amita Kamath, Christopher Clark, Zhen Zhu, Michal Shlapentokh-Rothman, and Jiasen Lu for several discussions that helped shape GRIT. Many thanks to Yuqun Wu for surface normal processing and experiments. Finally, we are grateful to Michal Guerquin, Jon Borchardt, M Kusold, and Michael Schmitz from the AI2 Reviz team for providing incredible web-tools and engineering support for GRIT. 

{\small
\bibliographystyle{ieee_fullname}
\bibliography{bib_derek}
}

\end{document}